# PICS: PIPELINE FOR IMAGE CAPTIONING AND SEARCH

Grant Rosario[1] and David Noever[2]

PeopleTec, 4901-D Corporate Drive, Huntsville, AL, USA, 35805
[1]grant.rosario@peopletec.com    [2] david.noever@peopletec.com

## ABSTRACT

*The growing volume of digital images necessitates advanced systems for efficient categorization and retrieval, presenting a significant challenge in database management and information retrieval. This paper introduces PICS (Pipeline for Image Captioning and Search), a novel approach designed to address the complexities inherent in organizing large-scale image repositories. PICS leverages the advancements in Large Language Models (LLMs) to automate the process of image captioning, offering a solution that transcends traditional manual annotation methods. The approach is rooted in the understanding that meaningful, AI-generated captions can significantly enhance the searchability and accessibility of images in large databases. By integrating sentiment analysis into the pipeline, PICS further enriches the metadata, enabling nuanced searches that extend beyond basic descriptors. This methodology not only simplifies the task of managing vast image collections but also sets a new precedent for accuracy and efficiency in image retrieval. The significance of PICS lies in its potential to transform image database systems, harnessing the power of machine learning and natural language processing to meet the demands of modern digital asset management.*

## KEYWORDS

*Image Captioning, Database Management, Large Language Models (LLMs), Image Retrieval*

## 1. INTRODUCTION

Digital images are becoming increasingly ubiquitous in our current age of tech devices, such as smartphones, computers, home security systems, smartwatches, etc., most of which contain some sort of camera. With this rise in actual image data, not to mention the rise in artificially generated image data from Vision-Language Models which are able to generate high-quality images from text, it's becoming increasingly important to store and organize image data in an efficient, repeatable way while also making sure the images are easily accessible. Achieving accurate image retrieval from a stored database presents many challenges due to the system needing to understand the details of the image in a way that is relevant to humans [Vijayaraju, Nivetha, 2019].

The concept of differentiating between the human interpretation of the image and what a computer algorithm can understand from pixel data is known as the semantic gap and bridging it is key to ensuring image retrieval systems can understand and respond to human queries [Kwaśnicka et al., 2018]. Additionally, image variation regarding lighting, perspective, occlusion, and quality also presents another significant challenge that can affect the performance of these systems making it important to create algorithms robust enough to handle these differing image styles [Jenicek et al., 2019; Zhao et al., 2020]. Another crucial challenge in image retrieval involves the standardization of annotations and metadata in which image metadata can frequently be incomplete, inconsistent, or absent altogether [Wang et al., 2009]. Lastly, integrating contextual information which goes beyond the actual image content, such as geographical data, temporal information, or in our case sentiment descriptors continues to be a pervasive challenge [Krojer et al, 2022]. Ultimately, image retrieval systems continue to attempt to solve these problems in order to balance accuracy, efficiency, and scalability in diverse real-world applications.

| | |
|---|---|
| 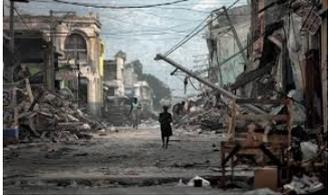 | 00afad9b-78f7-48d6-a44b-b046019a4548.jpg |
| 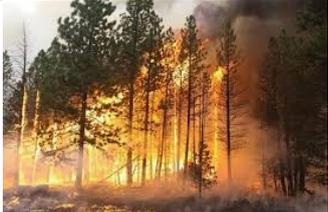 | 00bf1649-7f48-44ce-b208-a648023bd65a.jpg |
| 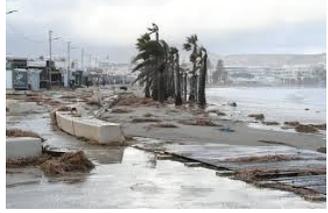 | 00c9dbac-4f02-4bdb-abf6-9dae2f4b8fc6.jpg |
| 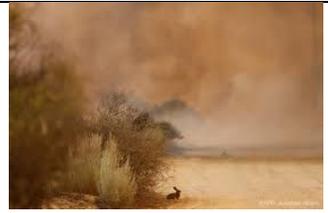 | 00c8305c-c779-4445-b3d6-ecfbce875ed5.jpg |

**Table 1: Left: Images in sentiment dataset. Right: Name of corresponding image.**

For the present work, we're introducing PICS, or "Pipeline for Image Captioning and Search", which seeks to address some of these challenges. In our implementation, we are utilizing the Image Sentiment Analysis dataset since it provides a large enough and contextually rich dataset of images as well as descriptive annotations we can attach for image retrieval [Hassan et al., 2022]. The novel steps of our pipeline consist of first addressing the naming convention issue of many image datasets. As shown in Table 1, most image datasets consist of either randomly generated image names or research-specific conventions that have no meaning to end users. The first step in PICS is to rename the images based on LLaVA and Mistral 7B by giving it a specific prompt and letting the model generate descriptive captions based on the image content [Liu et al., 2023; Jiang et al., 2023]. Secondly, we match the images to their corresponding human-generated sentiment annotations and add these annotations as metadata to each image. We then add these images to a database for image retrieval resulting in a database that we believe improves the current balance of Machine Learning based visual descriptions with diverse human-generated contextual annotations.

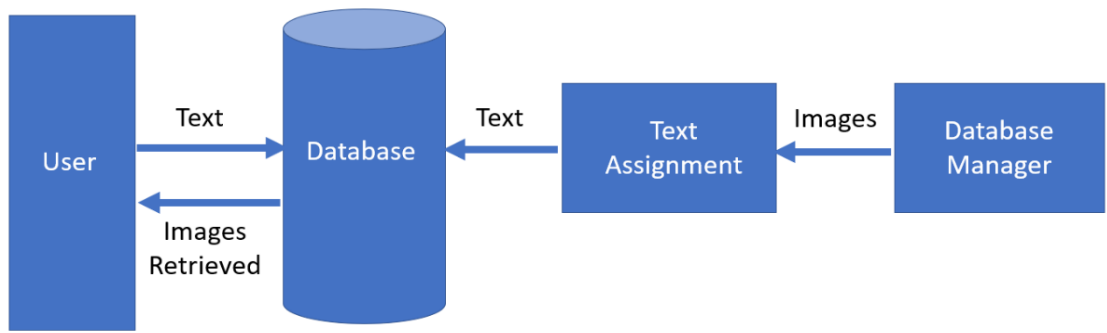

**Figure 1: Concept-Based Image Indexing Workflow [Baeldung]**

## 2. RELATED WORK

When it comes to image retrieval, there are two industry standard methods used today; Concept-Based Image Indexing and Content-Based Image Retreival [Chu et al., 2001; Ibtihaal et al., 2021].

### 2.1. Concept-Based Image Indexing

Sometimes referred to as text-based image retrieval, this method relies on indexing and retrieving images based on abstract concepts and semantics, typically provided by human labelers. Figure 1 shows the data and workflow structure of this method in action. The database manager populates the database with images and their corresponding textual annotations, labels, tags, etc., and the user can retrieve images by querying certain keywords, annotations, or descriptions that correspond with the residing labels. Concept-based image indexing returns any images that were assigned similar textual tags as what the user specified [Chu et al., 2001].

Early research in concept-based image indexing primarily focused on associating descriptive keywords or tags with images, either through manual annotation or using basic object recognition techniques. The breakthroughs in this period allowed users to search for images based on textual queries that matched these annotations. However, manually annotating images was and is labor-intensive and lacks scalability [Smeulders et al., 2000].

Later research sought to address the drawbacks of manual annotation by automating the workflow using unsupervised clustering methods [Jeon et al., 2003]. This was a promising step forward but many attempts at automation still required a base foundation of annotated images. However with the advent of machine learning and specifically the development of Convolutional Neural Networks (CNNs) made significant stride in the ease of labelling [Krizhevsky et al., 2012]. This paved the way for more sophisticated concept-based image indexing systems capable of understanding and categorizing image content at a near-human level.

Most recently, the field continues to evolve with research delving into more nuanced aspects of image understanding, such as sentiment analysis, applying experimental deep learning architectures, and utilizing LLMs to perform large scale image retrieval[Hongbin et al., 2022; Noh et al., 2017; Tan et al., 2022].

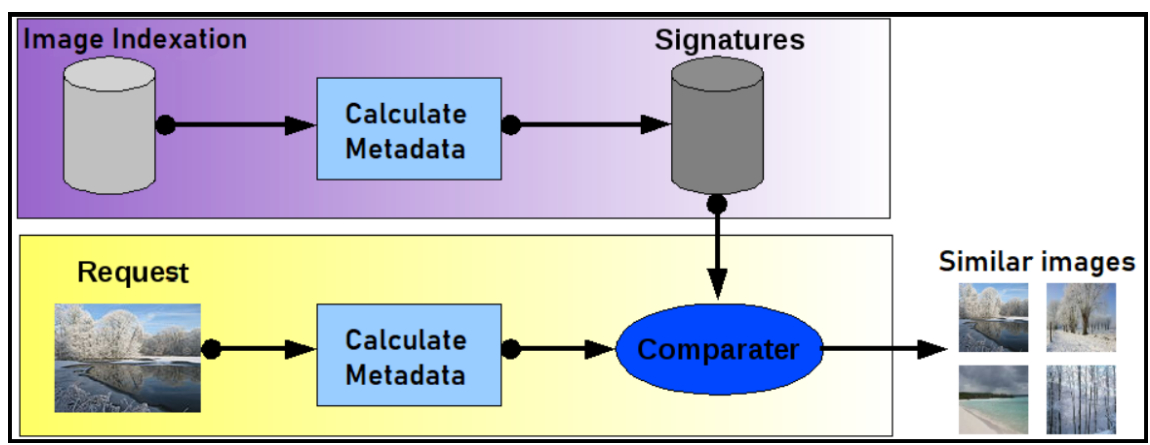

Figure 2: CBIR system schema [wikipedia]

## 2.2. Content-Based Image Retrieval (CBIR)

Content-based image retrieval (CBIR) referse to the process of retrieving relevant images from large databases based on the content present within them, such as colors, textures, shapes, or any other information that can be derived from the image itself. Figure 2 illustrates the typical schema of a CBIR system. The main obvious different from concept-based image indexing is that the user is making a request by providing a source image and asking for similar images rather than making a request based on text. The system then comapres the content of the query image with the conent of the database images. This makes this method unique in that it isn't reliant on manual annotation and text-based searching. CBIR systems aim to analyze and understand the visual content of images directly, which ideally allows for more intuitive and effective image searching [Smeulders et al., 2000].

Recent research in the field of CBIR has more or less surpassed the performance of concept-based image indexing. The advent of machine learning coupled with computer vision techniques enabled CBIR systems to visually understand images and significant components of them in ways strikingly similar to humans [Datta et al., 2008].

## 2. METHODS

The pipeline we're introducing, PICS, provides a unique outlook on the current image retrieval trends in that it combines some aspects of both concept-based image indexing and content-based image retrieval. As stated previously, to conduct our experiment, we are using the Sentiment

Analysis dataset due to it already containing a rich annotation dataset that corresponds with the images as well as contextually rich images that a Large Vision Model can potentially add more descriptive labels to.

| 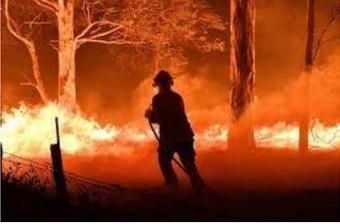 8c1aa7ed-1c6a-4cf2-8fas-d4a8dbbfdd3e.jpg | Prompt: The best description of the image in ten words or less is… | 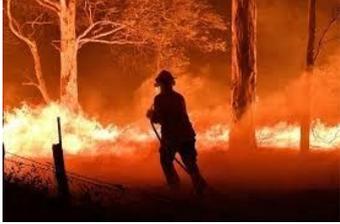 Fireman_with_hose_in_front_of_fire.jpg |
|---|---|---|

**Table 2: LLaVa prompt example. Left: Original image and name. Middle: LLaVa Prompt. Right: Resulting new image name/caption combo**

### 3.1. CBIR step

The first issue our pipeline addresses is the image naming convention problem. Most images that exist in datasets today are labelled with random conventions unknown to the end user, thereby causing a gap in the image data where there could be beneficial information. Our pipeline presents a solution to this problem by using a content-based machine learning approach. We begin by utilizing a bash script and llamafiles which enable users to turn a Large Language Models (LLMs) weights into executables, thereby enabling users to prompt LLMs just like they would an executable [Hood 2023; Tunney 2024]. The first step in this script is to use Mistral-7B to look at the image name and decide if it is already in readbale English or consists of non-readable letters, numbers, and/or symbols. If it is non-readable, we prompt the LLaVA Large Language and Vision Assistant llamafile to generate a new title for the image based on the content of the image itself. Table 2 shows the prompt we used and an example result. It's important to note that we save the original non-readable image name in the image metadata so that it's accessible when we need to look up the corresponding sentiment annotations.

### 3.2. Concept-based step

With this CBIR-like step done, our next step is taking the sentiment annotation extraction. This is what we would call the concept-based indexing step. We simply grab the original name from the image metadata and look up the corresponding sentiment annotation. Once we obtain the relevant sentiment tags based on the human-labeled sentiment annotations, we add these to the images as new metadata.

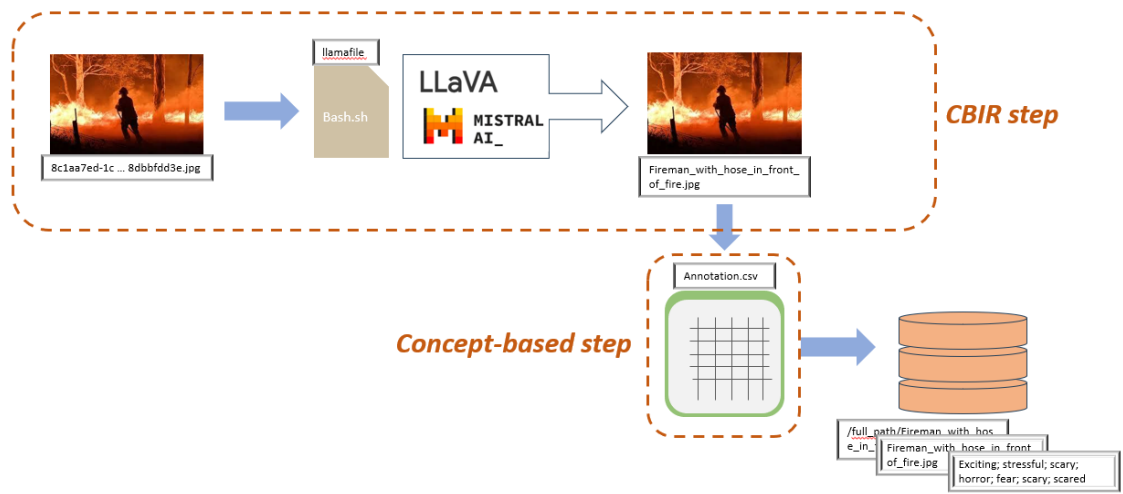

**Figure 3: PICS "Pipeline for Image Caption and Search"**

### 3.3. Database Creation

Lastly, we simply create a database with our image file paths, image names, and sentiment annotations. Figure 3 illustrates our schema workflow for PICS while keeping the source annotations agnostic, while we used sentiment annotations, any reliable annotations could be used.

## 3. RESULTS

The results of PICS provide a promising lookup relevance and accuracy. As shown in Table 3, typing in a keyword like happy and animal returns two highly relevant images to our search terms. We believe the most beneficial piece of PICS is the integration with current Large Vision Models. These models provide new avenues for creating highly accurate and relevant captions and image names in fraction of the time it would take human labelers to accomplish the same task.

| Search: Animal, happy | 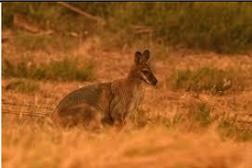 a_small_brown_animal_sitting_in_tall_ grass.jpg | 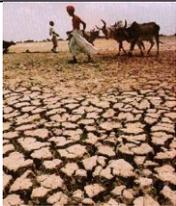 dry_land_with_animals _and_people_on_it.jpg |
|---|---|---|

## 4. DISCUSSION

We believe the PICS (Pipeline for Image Caption and Search) system represent a significant stride in automating the process of image annotation and retrieval. By integrating the capabilities of Large Language Models (LLMs) such as LLaVa and Mistral-7B, alongside human-generated labels from a sentiment analysis dataset, PICS has showcased the potential to substantially reduce the time and labor traditionally required for accurate image annotation and captioning. The usage of LLaVa and Mistral-7B has proven particularly promising, as these models not only understand the contextual nuances in images but also translate these perceptions into coherent and descriptive captions. This combination of perception and linguistic generation is what sets PICS apart from earlier systems that relied solely on fixed databases of keywords or ML-based pattern recognition.

An unforeseen advantage of PICS is its ability to process images at a speed that far surpasses manual labeling. The implications of this are multifold; it enables the rapid categorization of large datasets, facilitates the dynamic updating of image libraries, and supports real-time image retrieval applications.

However, the PICS system is not without its challenges. One of the primary concerns is the reliance on the quality and diversity of the training data used for the LLMs. Biases inherent in training datasets can lead to skewed or inaccurate captions, which necessitates careful curation of the training material. Additionally, the semantic gap—while narrowed—is still present. The current system excels in general contexts but may struggle with highly specialized or nuanced imagery that requires expert knowledge.

Looking forward, the integration of PICS with feedback loops where human inputs refine and correct the system's outputs could establish a synergistic relationship between AI-generated and human-generated annotations. This collaboration could lead to continuous improvement in the accuracy and relevance of the captions, while still retaining the efficiency of an automated system.

## ACKNOWLEDGMENTS

The authors thank the PeopleTec Technical Fellows program for its encouragement and project assistance.

**Authors**

| | |
|---|---|
| **Grant Rosario** has research experience in embedded applications and autonomous driving applications. He received his Masters from Florida Atlantic University in Computer Science and his Bachelors from Florida Gulf Coast University in Psychology. | 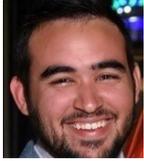 |
| **David Noever** has research experience with NASA and the Department of Defense in machine learning and data mining. He received his BS from Princeton University and his Ph.D. from Oxford University, as a Rhodes Scholar, in theoretical physics. | 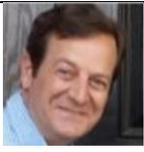 |